\documentclass[11pt,a4paper]{article}
\usepackage[hyperref]{acl2021}
\usepackage{times}
\usepackage{latexsym}
\usepackage{booktabs}

\usepackage{lipsum}
 \newcommand\blfootnote[1]{%
  \begingroup
  \renewcommand\thefootnote{}\footnote{#1}%
  \addtocounter{footnote}{-1}%
  \endgroup
}
\usepackage{enumerate}
\usepackage{graphicx}
\usepackage[inline]{enumitem}
\usepackage{microtype}

\aclfinalcopy 

\title{WLV-RIT at SemEval-2021 Task 5: A Neural Transformer Framework for Detecting Toxic Spans} 

\author{Tharindu Ranasinghe\textsuperscript{1}, Diptanu Sarkar\textsuperscript{2}, Marcos Zampieri\textsuperscript{2}, Alexander Ororbia\textsuperscript{2}\\
  \textsuperscript{1}University of Wolverhampton,UK \\
  \textsuperscript{2}Rochester Institute of Technology, USA\\
  \texttt{T.D.RanasingheHettiarachchige@wlv.ac.uk} 
  }

\date{}

\begin{document}
\maketitle
\begin{abstract} 
In recent years, the widespread use of social media has led to an increase in the generation of toxic and offensive content on online platforms. In response, social media platforms have worked on developing automatic detection methods and employing human moderators to cope with this deluge of offensive content. While various state-of-the-art statistical models have been applied to detect toxic posts, there are only a few studies that focus on detecting the words or expressions that make a post offensive. This motivates the organization of the SemEval-2021 Task 5: Toxic Spans Detection competition, which has provided participants with a dataset containing toxic spans annotation in English posts. In this paper, we present the WLV-RIT entry for the SemEval-2021 Task 5. Our best performing neural transformer model achieves an $0.68$ F1-Score. Furthermore, we develop an open-source framework for multilingual detection of offensive spans, i.e., MUDES, based on neural transformers that detect toxic spans in texts.
\end{abstract}

\blfootnote{WARNING: This paper contains text excerpts and words that are offensive in nature.}

\section{Introduction}
\label{sec:intro}

The widespread adoption and use of social media has led to a drastic increase in the generation of abusive and profane content on the web. To counter this deluge of negative content, social media companies and government institutions have turned to developing and applying computational models that can identify the various forms of offensive content online such as aggression \cite{kumar2018benchmarking,trac2020}, cyber-bullying \cite{rosa2019automatic}, and hate speech \cite{ridenhour2020detecting}. Prior work has either designed methods for identifying conversations that are likely to go awry \cite{zhang2018conversations,chang2020don} or  detecting offensive content and labelling posts at the instances level -- this has been the focus in the recent shared tasks like HASOC at FIRE 2019 \cite{mandl2019overview} and FIRE 2020 \cite{hasoc2020}, GermEval 2019 Task 2 \cite{strussoverview}, TRAC \cite{kumar2018benchmarking, trac2020}, HatEval \cite{basile2019semeval}, OffensEval at SemEval-2019 \cite{offenseval} and SemEval-2020 \cite{zampieri2020semeval}. 


With respect to identifying offensive language in conversations, comments, and posts, noticeable progress has been made with a variety of large, annotated datasets made available in recent years \cite{pitenis2020,SOLID}. The identification of the particular text spans that make a post offensive, however, has been mostly neglected \cite{mathew2020hatexplain} as current state-of-the-art offensive language identification models flag the entire post or comment but do not actually highlight the offensive parts. 
The pressing need for toxic span detection models to assist human content moderation, processing and flagging content in a more interpretable fashion, has motivated the organization of the SemEval-2021 Task 5: Toxic Spans Detection \cite{pav2020semeval}.



In this paper, we present the WLV-RIT submission to the SemEval-2021 Task 5. We explore several statistical learning models and report the performance of the best model, which is based on a neural transformer. Next, we generalise our approach to an open-source framework called MUDES: Multilingual Detection of Offensive Spans \cite{ranasinghemudes}. Alongside the framework, we also release the pre-trained models as well as a user-friendly web-based User Interface (UI) based on Docker, which provides the functionality of automatically identifying the offensive spans in a given input text.


\section{Related Work}
\label{sec:related}

\paragraph{Datasets} Over the past several years, multiple post-level, offensive language benchmark datasets have been released. In \citet{OLID}, the authors compiled an offensive language identification dataset with a three-layer hierarchical annotation scheme -- profanity, category, and target identification. \newcite{SOLID} further extended the dataset using a semi-supervised model that was trained with over nine million annotated English tweets. Recently, \newcite{mathew2020hatexplain} released the first benchmark dataset which covered the three primary areas of online hate-speech detection. The dataset contained a 3-class classification problem (hate-speech, offensive, or neither), a targeted community, as well as the spans that make the text hateful or offensive. Furthermore, offensive language datasets have been annotated in other languages such as Arabic \cite{mubarak2017abusive}, Danish \cite{sigurbergsson2020offensive}, Dutch \cite{tulkens2016dictionary}, French \cite{chiril-etal-2019-multilingual}, Greek \cite{pitenis2020}, Portuguese \cite{fortuna2019hierarchically}, Spanish \cite{basile-etal-2019-semeval}, and Turkish \cite{coltekin2020}.

Apart from the dataset released for SemEval-2021 Task 5, HateXplain \cite{mathew2020hatexplain} is, to the best of our knowledge, the only dataset that we could find that has been annotated at the word level. The dataset consists of $20,000$ posts from Gab and Twitter. Each data sample is annotated with one of the hate/offensive/normal labels, communities being targeted, and words of the text are marked by the annotators who support the label.

\paragraph{Models} In the past, trolling, aggression, and cyberbullying identification tasks on social media data have been approached using machine and deep learning-focused models \cite{kumar2018benchmarking}. Across several studies \cite{malmasi2017,malmasi2018challenges,waseem-hovy-2016-hateful} researchers have noted that $n$-gram based features are very useful when building reliable, automated hate-speech detection models. Statistical learning models aided with natural language processing (NLP) techniques are frequently used for post-level offensive and hateful language detection \cite{davidson2017automated, indurthi2019fermi}. Given the increased use of deep learning in NLP tasks, offensive language identification has seen the introduction of methods based on convolutional neural networks (CNNs) and Long Short-term Memory (LSTM) networks \cite{Badjatiya_2017, gamback-sikdar-2017-using, hettiarachchi-ranasinghe-2019-emoji}. The most common approach has been to use a word/character embedding model such as Word2vec \cite{NIPS2013_9aa42b31}, GloVe \cite{pennington2014glove}, or fastText \cite{mikolov2018advances} to embed words/tokens and then feed them to an artificial neural network (ANN) \cite{offenseval}.  

With the introduction of BERT \cite{devlin2019bert}, neural transformer models have become popular in offensive language identification. In hate speech and offensive content identification in Indo-European languages, the BERT model has been shown to outperform GRU (Gated Recurrent Unit) and LSTM-based models \cite{ranasinghe2019brums}. In \citet{hasoc2019}, the best performing teams on the task employed BERT-based pre-trained models that identified the type of hate and target of a (text) post. 

The SemEval-2019 Task 6 \cite{offenseval} presented the challenge of identifying and categorizing offensive posts on social media, which included three sub-tasks. In sub-task A: offensive language identification, \citet{liu2019nuli} applied a pre-trained BERT model to achieve the highest F1 score. In Sub-task B: automatic categorization of offense types, BERT-based models also achieved competitive rankings. We noticed similar trends in SemEval-2020 Task 12 \cite{zampieri2020semeval} as well. Not limited to English, transformer models have yielded strong results in resource-scarce languages like Bengali \cite{ranasinghe-etal-2020-multilingual} and Malayalam \cite{ranasinghe2020wlv} along with cross-lingual transfer learning from resource-rich languages \cite{ranasinghe-etal-2020-multilingual, ranasinghetallip}. Nonetheless, despite the recent success of statistical learning in offensive language detection problems, due to the lack of finer-grained, detailed datasets, models are limited in their ability to predict word-level labels.



\section{Task and Dataset}
\label{sec:task}

\begin{table*}[t]
\centering
\begin{tabular}{p{9cm}p{6cm}}
\toprule
\bf Post & \bf Offensive Spans  \\
\midrule
 \textcolor{red}{Stupid} hatcheries have completely \textcolor{red}{fucked} everything & [0, 1, 2, 3, 4, 5, 34, 35, 36, 37, 38, 39]  \\
 Victimitis: You are such an \textcolor{red}{asshole}. & [28, 29, 30, 31, 32, 33, 34]  \\
 So is his mother. They are silver spoon parasites. & []  \\
 You're just \textcolor{red}{silly}. & [12, 13, 14, 15, 16]  \\
\bottomrule
\end{tabular}
\caption{Four comments from the dataset along with their annotations. The offensive words are displayed in red and the spans are indicated by the character position in the instance.}
\label{tab:examples}
\end{table*}

In the SemEval-2021 Task 5 dataset, the sequence of words that makes a particular post or comment toxic is defined as a \emph{toxic span}. The dataset for this task is extracted from posts in the Civil Comments Dataset that have been found to be toxic. The practice dataset has $690$ instances out of which $43$ instances do not contain any toxic spans. The training dataset has a total of $7,939$ instances and comprises $485$ instances without any toxic spans. Each instance is composed of a list of toxic spans and the post (in English). In Table \ref{tab:examples}, we present four randomly selected examples from the training dataset along with their annotations.


\begin{figure}[ht]
\centering
\includegraphics[scale=0.78]{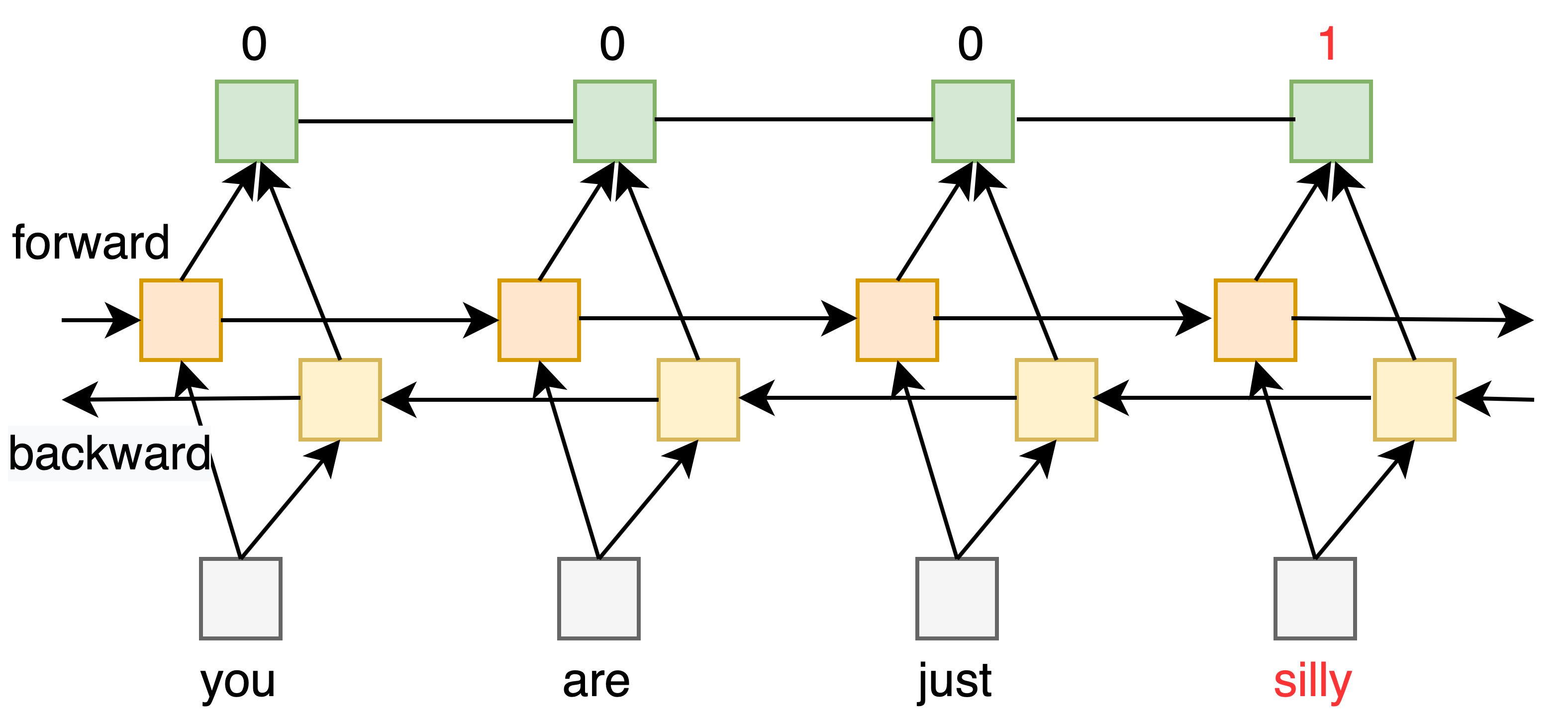}
\caption{The Bi-LSTM-CRF model. Green squares represent the top CRF layer.  Non-offensive and offensive tokens are shown as $0$ and $1$, respectively.}
\label{fig:lstm}
\end{figure}

\begin{figure*}[ht]
\centering
\includegraphics[scale=0.4]{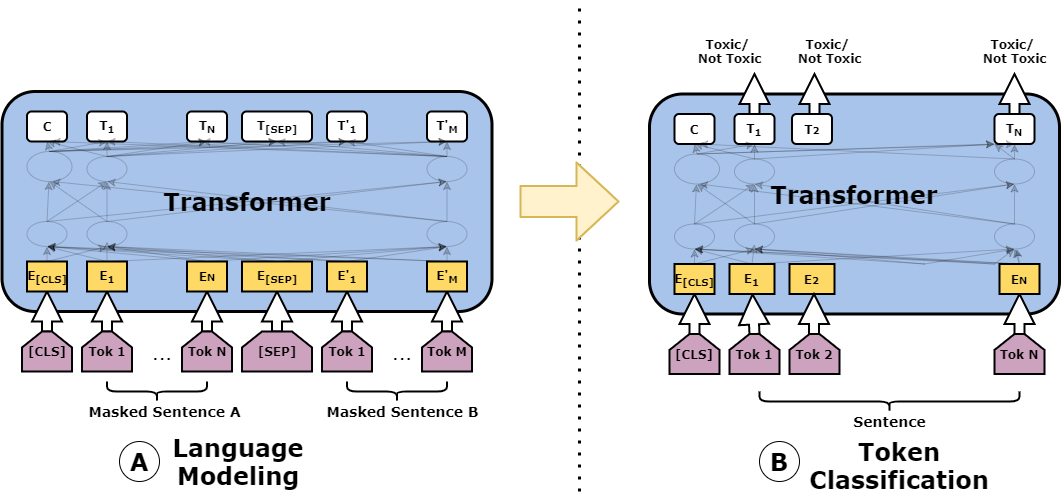}
\caption{The two-part model architecture. Part A depicts the language model and Part B is the token classifier. \cite{ranasinghemudes}}
\label{fig:architecture}
\end{figure*}

\section{Methodology}
\label{sec:method}

\subsection{Lexicon-based Word Match}
Lexicon-based word-matching algorithms often achieve balanced results. For the lexicon, we collected profanity words from online resources\footnote{\url{https://www.cs.cmu.edu/~biglou/resources/bad-words.txt}}\textsuperscript{,}\footnote{\url{https://github.com/RobertJGabriel/Google-profanity-words}}. Then, we added the toxic words present in the training dataset and we run a simple word matching algorithm the trie data structure. As anticipated, the algorithm does not evaluate the toxic spans contextually and misses censored swear words. For instance, the word \emph{f**k} is missed, which is not present in the lexicon. Nonetheless, this result provides as a useful baseline performance measurement for the task.

\subsection{Recurrent Networks: Long Short-Term Memory}
Long Short-term Memory (LSTM) is a recurrent neural network model that uses feedback connections to model temporal dependencies (past-to-present) in sequential data. Bidirectional LSTM (Bi-LSTM) is capable of learning contextual information both forwards and backwards in time compared to conventional LSTMs. In this study, we used the Bi-LSTM architecture given this bi-directional ability to model temporal dependencies. Conditional random fields (CRF) \cite{lafferty2001conditional} are a statistical model that are capable of incorporating context information and are highly used for sequence labeling tasks. A CRF connected to the top of the Bi-LSTM model provides a powerful way to model relationships between consecutive outputs (across time) and provides a means to efficiently utilize past and future tag information to predict the current tag. 

The final hybrid model is comparable to the previous state-of-the-art sequence tagging Bi-LSTM-CRF model \cite{DBLP:journals/corr/HuangXY15}. Figure \ref{fig:lstm} presents the Bi-LSTM-CRF architecture we designed for this study, which has 4.2 million trainable parameters. We trained the model on mini-batches of $16$ samples with a $0.005$ learning rate for $5$ epochs with a maximum sequence length of $200$.

\subsection{Neural Transformers}
Recently, pre-trained language models have been shown to be quite useful across a variety of NLP tasks, particularly those based on bidirectional neural transformers such as BERT \cite{devlin2019bert, 8975793}.  Transformer-based models have also been shown to be highly effective in sequence classification tasks such as named entity recognition (NER) \cite{luoma-pyysalo-2020-exploring}. In our work, we extend the BERT model by integrating a token level classifier. The token-level classifier is a linear transformation that takes the last hidden state of the sequence as the input and produces a label for each token as its output. In this case, each token will be predicted to have one of two possible labels -- toxic or not toxic. We fine-tuned the uncased BERT transformer model with a maximum sequence length of $400$ with batches of size of $16$. 

We also experimented with customising the layers in between the BERT transformer and token-classification layer by adding a CRF layer between them given that it has been shown that BERT-CRF architectures often outperform BERT baselines in similar sequence labeling tasks \cite{huang2019bertbased, souza2020portuguese}. Therefore, we added a sequential CRF layer on top of the BERT transformer and further incorporated dropout (probability of dropping a neuron was $0.2$) to introduce some regularization. 
Unfortunately, in our experiments, we found that adding a CRF layer does not significantly improve the final generalization results. 
Additionally, we experimented with transfer learning to identify if a further boost in model generalization was possible if we first trained a basic BERT transformer on HateXplain \cite{mathew2020hatexplain} and then fine-tuned it using our extended architecture as described above. However, the transfer learning process did not improve results any further.

\paragraph{Development of MUDES}
Given the success we observed using neural transformers such as BERT, we developed a (software) framework we call MUDES \cite{ranasinghemudes}: Multilingual Detection of Offensive Spans, an open-source framework based on transformers to detect toxic spans in texts. MUDES offers several capabilities in addition to the (automatic) token classification we described earlier. MUDES has the following components:
\begin {enumerate*} [label=\itshape\alph*\upshape)] 
\item \textbf{Language Modeler}: Fine-tuning transformer models using masked language modeling before performing the downstream task often leads to better results \cite{ranasinghe-hettiarachchi-2020-brums} and MUDES incorporates this, 
\item 
\textbf{Transformer Type Variety}: since there are many varieties of neural transformers, e.g., XLNet \cite{NEURIPS2019_dc6a7e65}, RoBERTa \cite{liu2019roberta} that have been shown to outperform BERT-based architectures \cite{ranasinghe-hettiarachchi-2020-brums, hettiarachchi-ranasinghe-2020-brums}, our software framework provides support for these architectures, and, finally, 
\item \textbf{Model Ensembling}:  multiple MUDES models with different random seeds can be trained and the final model prediction is the majority vote from all the models, aligning with the approach taken in \citet{hettiarachchi-ranasinghe-2020-infominer, hettiarachchi2021semeval, jauhiainen2021comparing}.
\end {enumerate*} 


The complete architecture of MUDES is depicted in Figure \ref{fig:architecture}. We used several popular transformer models including BERT \cite{devlin2019bert}, XLNET \cite{NEURIPS2019_dc6a7e65}, RoBERTa \cite{liu2019roberta}, SpanBERT \cite{joshi-etal-2020-spanbert}, and ALBERT \cite{Lan2020ALBERT:}. We compared these transformer architectures against the spaCy token classifier baseline (reported by the competition organisers) and report these results in Section \ref{sec:res}. Since adding a CRF layer did not improve the results in our models, we do not add this to MUDES.

Parameter optimization involved mini-batches of $8$ samples using the Adam update rule (global learning rate was $2\mathrm{e}{-5}$ and a linear warm-up schedule over $10$\% of the training data was used). Models were evaluated using a validation subset that contained $20$\% of the training data. Early stopping was executed if the validation loss did not improve over $10$ evaluation steps. Models were trained for $3$ epochs on an Nvidia Tesla K80 GPU using only the training set provided. 

\section{Evaluation and Results}
\label{sec:res}

For evaluation, we followed the same procedure that the task organisers have used to evaluate the systems. 

Let system $A_i$ return a set $S^t_{A_i}$ of character offsets for parts of a text post that have been found to be toxic. Let $G_t$ be the character offsets of the ground truth annotations of $t$. We compute the F1 score of system $A_i$ with respect to the ground truth $G$ for post $t$ as mentioned in Equation \ref{equation_f1} where $\vert$ ·$\vert$ denotes set cardinality. $P^{t}$ and $R^{t}$ measure the precision and recall, respectively.

\begin{equation}
\label{equation_f1}
F_{1}^{t}\left(A_{i}, G\right)=\frac{2 \cdot P^{t}\left(A_{i}, G\right) \cdot R^{t}\left(A_{i}, G\right)}{P^{t}\left(A_{i}, G\right)+R^{t}\left(A_{i}, G\right)}
\end{equation}

\vspace{2mm}

\begin{table}[!ht]
\begin{center}
\scalebox{0.98}{
\begin{tabular}{ c|c|c } 
 \hline
 \textbf{Model} & \textbf{Trial F1} & \textbf{Test F1}\\ 
 \hline
   MUDES RoBERTa & 0.6886 & 0.6801\\
   MUDES BERT & 0.6771 & 0.6698  \\
   MUDES SPANBert & 0.6751 & 0.6675  \\
   MUDES XLNet & 0.6722 & 0.6653  \\
   BERT & 0.6738 & 0.6538 \\
   BERT-CRF & 0.6643 & 0.6517 \\
   BERT HateXplain & 0.6387 & 0.6326 \\
   spaCy baseline & 0.5976 & 0.5976\\
   Bi-LSTM-CRF & 0.5631 & 0.5398 \\
   Lexicon word match & 0.3378 & 0.4086 \\
 \hline
\end{tabular}
}
\end{center}
\caption{Results ordered by test F1 score. The Trial F1 column shows the F1 scores on the trial set and the Test F1 column shows the F1 scores for test set.} 
\label{tab:results}
\end{table}

\noindent Observe in Table \ref{tab:results} that all of our deep neural-based models outperformed the spaCy baseline while the lexicon-based word match algorithm provided fairly good results despite it being an unsupervised method. Our best model is the MUDES RoBERTa model which scored $0.68$ F1 score in the test set and is very compatible with the $0.70$ F1 score that the best model scored in the competition. Furthermore, it is clear that the additional features supported by our MUDES framework, e.g., language modeling and ensembling, improves the results over a vanilla BERT transformer. 

\section{Conclusion and Future Work}
\label{sec:conc}
In this paper, we presented the WLV-RIT approach for tackling the SemEval-2021 Task 5: Toxic Spans Detection. SemEval-2021 Task 5 provided participants with the opportunity of testing computational models to identify token spans in toxic posts as opposed to previous related SemEval tasks such as HatEval and OffensEval that provided participants with datasets annotated at the instance level. We believe that word-level predictions are an important step towards explainable offensive language identification.

We experimented with several methods including a lexicon-based word match, LSTMs, and neural transformers. Our results demonstrated that transformer models offered the best generalization results and, given the success observed, we developed MUDES, an open-source software framework based on neural transformers focused on detecting toxic spans in texts. 
With MUDES. we release two English models that performed best for this task \cite{ranasinghemudes}. A large model; en-large based on roberta-large which is more accurate, but has a low efficiency regarding space and time. The base model based on xlnet-base-cased; en-base is efficient, but has a comparatively low accuracy than the en-large model. All pre-trained models are available on Hugging Face Model Hub \cite{wolf-etal-2020-transformers}\footnote{Available on \url{https://huggingface.co/mudes}}.
We also make MUDES available as a Python package\footnote{Available at \url{https://pypi.org/project/mudes/}} and set up as an open-source project\footnote{The MUDES GitHub repository is available at \url{https://github.com/tharindudr/MUDES}}. In addition, a prototype User Interface (UI) of MUDES has been made accessible to the general public\footnote{The UI can be accessed from \url{http://rgcl.wlv.ac.uk/mudes/}} based on Docker\footnote{Available at \url{https://hub.docker.com/r/tharindudr/mudes}}.
 

In terms of future work, we would like to experiment with multi-task (neural) architectures that can be used for offensive language identification capable of carrying out predictions at both the word-level and post-level jointly. Furthermore, we would like to evaluate multi-task architectures on multi-domain and multilingual settings as well as broaden our experimental comparison to other types of recurrent network models, such as the Delta-RNN \cite{ororbia2017learning}.


\section*{Acknowledgments}

We would like to thank the shared task organizers for making this interesting dataset available. We further thank the anonymous SemEval reviewers for their insightful feedback.

\bibliography{offensive}

\begin{thebibliography}{60}
\expandafter\ifx\csname natexlab\endcsname\relax\def\natexlab#1{#1}\fi

\bibitem[{Badjatiya et~al.(2017)Badjatiya, Gupta, Gupta, and
  Varma}]{Badjatiya_2017}
Pinkesh Badjatiya, Shashank Gupta, Manish Gupta, and Vasudeva Varma. 2017.
\newblock Deep learning for hate speech detection in tweets.
\newblock In \emph{Proceedings of WWW}.

\bibitem[{Basile et~al.(2019{\natexlab{a}})Basile, Bosco, Fersini, Nozza,
  Patti, Pardo, Rosso, and Sanguinetti}]{basile2019semeval}
Valerio Basile, Cristina Bosco, Elisabetta Fersini, Debora Nozza, Viviana
  Patti, Francisco Manuel~Rangel Pardo, Paolo Rosso, and Manuela Sanguinetti.
  2019{\natexlab{a}}.
\newblock {Semeval-2019 task 5: Multilingual detection of hate speech against
  immigrants and women in twitter}.
\newblock In \emph{Proceedings of SemEval}.

\bibitem[{Basile et~al.(2019{\natexlab{b}})Basile, Bosco, Fersini, Nozza,
  Patti, Rangel~Pardo, Rosso, and Sanguinetti}]{basile-etal-2019-semeval}
Valerio Basile, Cristina Bosco, Elisabetta Fersini, Debora Nozza, Viviana
  Patti, Francisco~Manuel Rangel~Pardo, Paolo Rosso, and Manuela Sanguinetti.
  2019{\natexlab{b}}.
\newblock {S}em{E}val-2019 task 5: Multilingual detection of hate speech
  against immigrants and women in {T}witter.
\newblock In \emph{Proceedings of SemEval}.

\bibitem[{\c{C}\"{o}ltekin(2020)}]{coltekin2020}
\c{C}a\u{g}r{\i} \c{C}\"{o}ltekin. 2020.
\newblock {A Corpus of Turkish Offensive Language on Social Media}.
\newblock In \emph{Proceedings of LREC}.

\bibitem[{Chang et~al.(2020)Chang, Cheng, and
  Danescu-Niculescu-Mizil}]{chang2020don}
Jonathan~P Chang, Justin Cheng, and Cristian Danescu-Niculescu-Mizil. 2020.
\newblock Don’t let me be misunderstood: Comparing intentions and perceptions
  in online discussions.
\newblock In \emph{Proceedings of WWW}.

\bibitem[{Chiril et~al.(2019)Chiril, Benamara~Zitoune, Moriceau, Coulomb-Gully,
  and Kumar}]{chiril-etal-2019-multilingual}
Patricia Chiril, Farah Benamara~Zitoune, V{\'e}ronique Moriceau, Marl{\`e}ne
  Coulomb-Gully, and Abhishek Kumar. 2019.
\newblock Multilingual and multitarget hate speech detection in tweets.
\newblock In \emph{Proceedings of TALN}.

\bibitem[{Davidson et~al.(2017)Davidson, Warmsley, Macy, and
  Weber}]{davidson2017automated}
Thomas Davidson, Dana Warmsley, Michael Macy, and Ingmar Weber. 2017.
\newblock {Automated Hate Speech Detection and the Problem of Offensive
  Language}.
\newblock In \emph{Proceedings of ICWSM}.

\bibitem[{Devlin et~al.(2019)Devlin, Chang, Lee, and
  Toutanova}]{devlin2019bert}
Jacob Devlin, Ming-Wei Chang, Kenton Lee, and Kristina Toutanova. 2019.
\newblock {BERT: Pre-training of Deep Bidirectional Transformers for Language
  Understanding}.
\newblock In \emph{Proceedings of NAACL}.

\bibitem[{Fortuna et~al.(2019)Fortuna, da~Silva, Wanner, Nunes
  et~al.}]{fortuna2019hierarchically}
Paula Fortuna, Joao~Rocha da~Silva, Leo Wanner, S{\'e}rgio Nunes, et~al. 2019.
\newblock {A Hierarchically-labeled Portuguese Hate Speech Dataset}.
\newblock In \emph{Proceedings of ALW}.

\bibitem[{Gamb{\"a}ck and Sikdar(2017)}]{gamback-sikdar-2017-using}
Bj{\"o}rn Gamb{\"a}ck and Utpal~Kumar Sikdar. 2017.
\newblock Using convolutional neural networks to classify hate-speech.
\newblock In \emph{Proceedings of ALW}.

\bibitem[{Hettiarachchi and
  Ranasinghe(2019)}]{hettiarachchi-ranasinghe-2019-emoji}
Hansi Hettiarachchi and Tharindu Ranasinghe. 2019.
\newblock Emoji powered capsule network to detect type and target of offensive
  posts in social media.
\newblock In \emph{Proceedings of RANLP}.

\bibitem[{Hettiarachchi and
  Ranasinghe(2020{\natexlab{a}})}]{hettiarachchi-ranasinghe-2020-brums}
Hansi Hettiarachchi and Tharindu Ranasinghe. 2020{\natexlab{a}}.
\newblock {BRUMS} at {S}em{E}val-2020 task 3: Contextualised embeddings for
  predicting the (graded) effect of context in word similarity.
\newblock In \emph{Proceedings of SemEval}.

\bibitem[{Hettiarachchi and
  Ranasinghe(2020{\natexlab{b}})}]{hettiarachchi-ranasinghe-2020-infominer}
Hansi Hettiarachchi and Tharindu Ranasinghe. 2020{\natexlab{b}}.
\newblock {I}nfo{M}iner at {WNUT}-2020 task 2: Transformer-based covid-19
  informative tweet extraction.
\newblock In \emph{Proceedings of W-NUT}.

\bibitem[{Hettiarachchi and Ranasinghe(2021)}]{hettiarachchi2021semeval}
Hansi Hettiarachchi and Tharindu Ranasinghe. 2021.
\newblock {TransWiC at SemEval-2021 Task 2: Transformer-based Multilingual and
  Cross-lingual Word-in-Context Disambiguation}.
\newblock In \emph{Proceedings of SemEval}.

\bibitem[{Huang et~al.(2019)Huang, Cheng, Wang, and Chu}]{huang2019bertbased}
Weipeng Huang, Xingyi Cheng, Taifeng Wang, and Wei Chu. 2019.
\newblock {BERT-Based Multi-Head Selection for Joint Entity-Relation
  Extraction}.
\newblock In \emph{Proceedings of NLPCC}.

\bibitem[{Huang et~al.(2015)Huang, Xu, and Yu}]{DBLP:journals/corr/HuangXY15}
Zhiheng Huang, Wei Xu, and Kai Yu. 2015.
\newblock Bidirectional {LSTM-CRF} models for sequence tagging.
\newblock \emph{arXiv preprint arXiv:1508.01991}.

\bibitem[{Indurthi et~al.(2019)Indurthi, Syed, Shrivastava, Chakravartula,
  Gupta, and Varma}]{indurthi2019fermi}
Vijayasaradhi Indurthi, Bakhtiyar Syed, Manish Shrivastava, Nikhil
  Chakravartula, Manish Gupta, and Vasudeva Varma. 2019.
\newblock {FERMI} at {S}em{E}val-2019 task 5: Using sentence embeddings to
  identify hate speech against immigrants and women in {T}witter.
\newblock In \emph{Proceedings of SemEval}.

\bibitem[{Jauhiainen et~al.(2021)Jauhiainen, Ranasinghe, and
  Zampieri}]{jauhiainen2021comparing}
Tommi Jauhiainen, Tharindu Ranasinghe, and Marcos Zampieri. 2021.
\newblock Comparing approaches to dravidian language identification.
\newblock In \emph{Proceedings of VarDial}.

\bibitem[{Joshi et~al.(2020)Joshi, Chen, Liu, Weld, Zettlemoyer, and
  Levy}]{joshi-etal-2020-spanbert}
Mandar Joshi, Danqi Chen, Yinhan Liu, Daniel~S. Weld, Luke Zettlemoyer, and
  Omer Levy. 2020.
\newblock {S}pan{BERT}: Improving pre-training by representing and predicting
  spans.
\newblock \emph{Proceedings of TACL}.

\bibitem[{Kumar et~al.(2018)Kumar, Ojha, Malmasi, and
  Zampieri}]{kumar2018benchmarking}
Ritesh Kumar, Atul~Kr Ojha, Shervin Malmasi, and Marcos Zampieri. 2018.
\newblock Benchmarking aggression identification in social media.
\newblock In \emph{Proceedings of TRAC}.

\bibitem[{Kumar et~al.(2020)Kumar, Ojha, Malmasi, and Zampieri}]{trac2020}
Ritesh Kumar, Atul~Kr. Ojha, Shervin Malmasi, and Marcos Zampieri. 2020.
\newblock {Evaluating Aggression Identification in Social Media}.
\newblock In \emph{Proceedings of TRAC}.

\bibitem[{Lafferty et~al.(2001)Lafferty, McCallum, and
  Pereira}]{lafferty2001conditional}
John~D. Lafferty, Andrew McCallum, and Fernando C.~N. Pereira. 2001.
\newblock Conditional random fields: Probabilistic models for segmenting and
  labeling sequence data.
\newblock In \emph{Proceedings of ICML}.

\bibitem[{Lan et~al.(2020)Lan, Chen, Goodman, Gimpel, Sharma, and
  Soricut}]{Lan2020ALBERT:}
Zhenzhong Lan, Mingda Chen, Sebastian Goodman, Kevin Gimpel, Piyush Sharma, and
  Radu Soricut. 2020.
\newblock {ALBERT: A Lite BERT for Self-supervised Learning of Language
  Representations}.
\newblock In \emph{Proceedings of ICLR}.

\bibitem[{{Li} et~al.(2019){Li}, {Gao}, {Zhou}, {Huang}, {Zhang}, and
  {Li}}]{8975793}
W.~{Li}, S.~{Gao}, H.~{Zhou}, Z.~{Huang}, K.~{Zhang}, and W.~{Li}. 2019.
\newblock {The Automatic Text Classification Method Based on BERT and Feature
  Union}.
\newblock In \emph{Proceedings of ICPADS}.

\bibitem[{Liu et~al.(2019{\natexlab{a}})Liu, Li, and Zou}]{liu2019nuli}
Ping Liu, Wen Li, and Liang Zou. 2019{\natexlab{a}}.
\newblock {NULI} at {S}em{E}val-2019 task 6: Transfer learning for offensive
  language detection using bidirectional transformers.
\newblock In \emph{Proceedings of SemEval}.

\bibitem[{Liu et~al.(2019{\natexlab{b}})Liu, Ott, Goyal, Du, Joshi, Chen, Levy,
  Lewis, Zettlemoyer, and Stoyanov}]{liu2019roberta}
Yinhan Liu, Myle Ott, Naman Goyal, Jingfei Du, Mandar Joshi, Danqi Chen, Omer
  Levy, Mike Lewis, Luke Zettlemoyer, and Veselin Stoyanov. 2019{\natexlab{b}}.
\newblock {RoBERTa: A Robustly Optimized BERT Pretraining Approach}.
\newblock \emph{arXiv preprint arXiv:1907.11692}.

\bibitem[{Luoma and Pyysalo(2020)}]{luoma-pyysalo-2020-exploring}
Jouni Luoma and Sampo Pyysalo. 2020.
\newblock {Exploring Cross-sentence Contexts for Named Entity Recognition with
  {BERT}}.
\newblock In \emph{Proceedings of COLING}.

\bibitem[{Malmasi and Zampieri(2017)}]{malmasi2017}
Shervin Malmasi and Marcos Zampieri. 2017.
\newblock {Detecting Hate Speech in Social Media}.
\newblock In \emph{Proceedings of RANLP}.

\bibitem[{Malmasi and Zampieri(2018)}]{malmasi2018challenges}
Shervin Malmasi and Marcos Zampieri. 2018.
\newblock {Challenges in Discriminating Profanity from Hate Speech}.
\newblock \emph{Journal of Experimental \& Theoretical Artificial
  Intelligence}, 30:1--16.

\bibitem[{Mandl et~al.(2020)Mandl, Modha, Kumar~M, and
  Chakravarthi}]{hasoc2020}
Thomas Mandl, Sandip Modha, Anand Kumar~M, and Bharathi~Raja Chakravarthi.
  2020.
\newblock Overview of the hasoc track at fire 2020: Hate speech and offensive
  language identification in tamil, malayalam, hindi, english and german.
\newblock In \emph{Proceedings of FIRE}.

\bibitem[{Mandl et~al.(2019{\natexlab{a}})Mandl, Modha, Majumder, Patel, Dave,
  Mandlia, and Patel}]{mandl2019overview}
Thomas Mandl, Sandip Modha, Prasenjit Majumder, Daksh Patel, Mohana Dave,
  Chintak Mandlia, and Aditya Patel. 2019{\natexlab{a}}.
\newblock {Overview of the hasoc track at fire 2019: Hate speech and offensive
  content identification in indo-european languages}.
\newblock In \emph{Proceedings of FIRE}.

\bibitem[{Mandl et~al.(2019{\natexlab{b}})Mandl, Modha, Majumder, Patel, Dave,
  Mandlia, and Patel}]{hasoc2019}
Thomas Mandl, Sandip Modha, Prasenjit Majumder, Daksh Patel, Mohana Dave,
  Chintak Mandlia, and Aditya Patel. 2019{\natexlab{b}}.
\newblock Overview of the hasoc track at fire 2019: Hate speech and offensive
  content identification in indo-european languages.
\newblock In \emph{Proceedings of FIRE}.

\bibitem[{Mathew et~al.(2021)Mathew, Saha, Yimam, Biemann, Goyal, and
  Mukherjee}]{mathew2020hatexplain}
Binny Mathew, Punyajoy Saha, Seid~Muhie Yimam, Chris Biemann, Pawan Goyal, and
  Animesh Mukherjee. 2021.
\newblock Hatexplain: A benchmark dataset for explainable hate speech
  detection.
\newblock In \emph{Proceedings of AAAI}.

\bibitem[{Mikolov et~al.(2018)Mikolov, Grave, Bojanowski, Puhrsch, and
  Joulin}]{mikolov2018advances}
Tomas Mikolov, Edouard Grave, Piotr Bojanowski, Christian Puhrsch, and Armand
  Joulin. 2018.
\newblock Advances in pre-training distributed word representations.
\newblock In \emph{Proceedings of LREC}.

\bibitem[{Mikolov et~al.(2013)Mikolov, Sutskever, Chen, Corrado, and
  Dean}]{NIPS2013_9aa42b31}
Tomas Mikolov, Ilya Sutskever, Kai Chen, Greg~S Corrado, and Jeff Dean. 2013.
\newblock Distributed representations of words and phrases and their
  compositionality.
\newblock In \emph{Proceedings of NeurIPS}.

\bibitem[{Mubarak et~al.(2017)Mubarak, Darwish, and Magdy}]{mubarak2017abusive}
Hamdy Mubarak, Kareem Darwish, and Walid Magdy. 2017.
\newblock Abusive language detection on {A}rabic social media.
\newblock In \emph{Proceedings of ALW}.

\bibitem[{Ororbia~II et~al.(2017)Ororbia~II, Mikolov, and
  Reitter}]{ororbia2017learning}
Alexander~G Ororbia~II, Tomas Mikolov, and David Reitter. 2017.
\newblock Learning simpler language models with the differential state
  framework.
\newblock \emph{Neural computation}, 29(12):3327--3352.

\bibitem[{Pavlopoulos et~al.(2021)Pavlopoulos, Laugier, Sorensen, and
  Androutsopoulos}]{pav2020semeval}
John Pavlopoulos, Léo Laugier, Jeffrey Sorensen, and Ion Androutsopoulos.
  2021.
\newblock Semeval-2021 task 5: Toxic spans detection.
\newblock In \emph{Proceedings of SemEval}.

\bibitem[{Pennington et~al.(2014)Pennington, Socher, and
  Manning}]{pennington2014glove}
Jeffrey Pennington, Richard Socher, and Christopher~D. Manning. 2014.
\newblock {GloVe: Global Vectors for Word Representation}.
\newblock In \emph{Proceedings of EMNLP}.

\bibitem[{Pitenis et~al.(2020)Pitenis, Zampieri, and Ranasinghe}]{pitenis2020}
Zeses Pitenis, Marcos Zampieri, and Tharindu Ranasinghe. 2020.
\newblock {Offensive Language Identification in Greek}.
\newblock In \emph{Proceedings of LREC}.

\bibitem[{Ranasinghe et~al.(2020)Ranasinghe, Gupte, Zampieri, and
  Nwogu}]{ranasinghe2020wlv}
Tharindu Ranasinghe, Sarthak Gupte, Marcos Zampieri, and Ifeoma Nwogu. 2020.
\newblock {WLV-RIT at HASOC-Dravidian-CodeMix-FIRE2020: Offensive Language
  Identification in Code-switched YouTube Comments}.
\newblock In \emph{Proceedings of FIRE}.

\bibitem[{Ranasinghe and
  Hettiarachchi(2020)}]{ranasinghe-hettiarachchi-2020-brums}
Tharindu Ranasinghe and Hansi Hettiarachchi. 2020.
\newblock {BRUMS} at {S}em{E}val-2020 task 12: Transformer based multilingual
  offensive language identification in social media.
\newblock In \emph{Proceedings of SemEval}.

\bibitem[{Ranasinghe and Zampieri(2020)}]{ranasinghe-etal-2020-multilingual}
Tharindu Ranasinghe and Marcos Zampieri. 2020.
\newblock {{Multilingual Offensive Language Identification with Cross-lingual
  Embeddings}}.
\newblock In \emph{Proceedings of EMNLP}.

\bibitem[{Ranasinghe and Zampieri(2021{\natexlab{a}})}]{ranasinghemudes}
Tharindu Ranasinghe and Marcos Zampieri. 2021{\natexlab{a}}.
\newblock {MUDES: Multilingual Detection of Offensive Spans}.
\newblock In \emph{Proceedings of NAACL}.

\bibitem[{Ranasinghe and Zampieri(2021{\natexlab{b}})}]{ranasinghetallip}
Tharindu Ranasinghe and Marcos Zampieri. 2021{\natexlab{b}}.
\newblock {Multilingual Offensive Language Identification for Low-resource
  Languages}.
\newblock \emph{ACM Transactions on Asian and Low-Resource Language Information
  Processing (TALLIP)}.

\bibitem[{Ranasinghe et~al.(2019)Ranasinghe, Zampieri, and
  Hettiarachchi}]{ranasinghe2019brums}
Tharindu Ranasinghe, Marcos Zampieri, and Hansi Hettiarachchi. 2019.
\newblock {BRUMS at HASOC 2019: Deep Learning Models for Multilingual Hate
  Speech and Offensive Language Identification}.
\newblock In \emph{Proceedings of FIRE}.

\bibitem[{Ridenhour et~al.(2020)Ridenhour, Bagavathi, Raisi, and
  Krishnan}]{ridenhour2020detecting}
Michael Ridenhour, Arunkumar Bagavathi, Elaheh Raisi, and Siddharth Krishnan.
  2020.
\newblock {Detecting Online Hate Speech: Approaches Using Weak Supervision and
  Network Embedding Models}.
\newblock In \emph{Proceedings of SBP-BRiMS}.

\bibitem[{Rosa et~al.(2019)Rosa, Pereira, Ribeiro, Ferreira, Carvalho,
  Oliveira, Coheur, Paulino, Sim{\~a}o, and Trancoso}]{rosa2019automatic}
Hugo Rosa, N~Pereira, Ricardo Ribeiro, Paula~Costa Ferreira, Joao~Paulo
  Carvalho, S~Oliveira, Lu{\'\i}sa Coheur, Paula Paulino, AM~Veiga Sim{\~a}o,
  and Isabel Trancoso. 2019.
\newblock Automatic cyberbullying detection: A systematic review.
\newblock \emph{Computers in Human Behavior}, 93:333--345.

\bibitem[{Rosenthal et~al.(2020)Rosenthal, Atanasova, Karadzhov, Zampieri, and
  Nakov}]{SOLID}
Sara Rosenthal, Pepa Atanasova, Georgi Karadzhov, Marcos Zampieri, and Preslav
  Nakov. 2020.
\newblock A large-scale semi-supervised dataset for offensive language
  identification.
\newblock \emph{arXiv preprint arXiv:2004.14454}.

\bibitem[{Sigurbergsson and Derczynski(2020)}]{sigurbergsson2020offensive}
Gudbjartur~Ingi Sigurbergsson and Leon Derczynski. 2020.
\newblock {Offensive Language and Hate Speech Detection for Danish}.
\newblock In \emph{Proceedings of LREC}.

\bibitem[{Souza et~al.(2020)Souza, Nogueira, and Lotufo}]{souza2020portuguese}
Fábio Souza, Rodrigo Nogueira, and Roberto Lotufo. 2020.
\newblock {Portuguese Named Entity Recognition using BERT-CRF}.
\newblock \emph{arXiv preprint arXiv:1909.10649}.

\bibitem[{Stru{\ss} et~al.(2019)Stru{\ss}, Siegel, Ruppenhofer, Wiegand,
  ScienceCampus, and Klenner}]{strussoverview}
Julia~Maria Stru{\ss}, Melanie Siegel, Josef Ruppenhofer, Michael Wiegand,
  Leibniz ScienceCampus, and Manfred Klenner. 2019.
\newblock Overview of germeval task 2, 2019 shared task on the identification
  of offensive language.
\newblock In \emph{Proceedings of KONVENS}.

\bibitem[{Tulkens et~al.(2016)Tulkens, Hilte, Lodewyckx, Verhoeven, and
  Daelemans}]{tulkens2016dictionary}
St{\'e}phan Tulkens, Lisa Hilte, Elise Lodewyckx, Ben Verhoeven, and Walter
  Daelemans. 2016.
\newblock {A Dictionary-based Approach to Racism Detection in Dutch Social
  Media}.
\newblock In \emph{Proceedings of TA-COS}.

\bibitem[{Waseem and Hovy(2016)}]{waseem-hovy-2016-hateful}
Zeerak Waseem and Dirk Hovy. 2016.
\newblock Hateful symbols or hateful people? predictive features for hate
  speech detection on {T}witter.
\newblock In \emph{Proceedings of NAACL Student Research Workshop}.

\bibitem[{Wolf et~al.(2020)Wolf, Debut, Sanh, Chaumond, Delangue, Moi, Cistac,
  Rault, Louf, Funtowicz, Davison, Shleifer, von Platen, Ma, Jernite, Plu, Xu,
  Scao, Gugger, Drame, Lhoest, and Rush}]{wolf-etal-2020-transformers}
Thomas Wolf, Lysandre Debut, Victor Sanh, Julien Chaumond, Clement Delangue,
  Anthony Moi, Pierric Cistac, Tim Rault, Rémi Louf, Morgan Funtowicz, Joe
  Davison, Sam Shleifer, Patrick von Platen, Clara Ma, Yacine Jernite, Julien
  Plu, Canwen Xu, Teven~Le Scao, Sylvain Gugger, Mariama Drame, Quentin Lhoest,
  and Alexander~M. Rush. 2020.
\newblock Transformers: State-of-the-art natural language processing.
\newblock In \emph{Proceedings of EMNLP}.

\bibitem[{Yang et~al.(2019)Yang, Dai, Yang, Carbonell, Salakhutdinov, and
  Le}]{NEURIPS2019_dc6a7e65}
Zhilin Yang, Zihang Dai, Yiming Yang, Jaime Carbonell, Russ~R Salakhutdinov,
  and Quoc~V Le. 2019.
\newblock {XLNet: Generalized Autoregressive Pretraining for Language
  Understanding}.
\newblock In \emph{Proceedings of NeurIPS}.

\bibitem[{Zampieri et~al.(2019{\natexlab{a}})Zampieri, Malmasi, Nakov,
  Rosenthal, Farra, and Kumar}]{OLID}
Marcos Zampieri, Shervin Malmasi, Preslav Nakov, Sara Rosenthal, Noura Farra,
  and Ritesh Kumar. 2019{\natexlab{a}}.
\newblock Predicting the type and target of offensive posts in social media.
\newblock In \emph{Proceedings of NAACL}.

\bibitem[{Zampieri et~al.(2019{\natexlab{b}})Zampieri, Malmasi, Nakov,
  Rosenthal, Farra, and Kumar}]{offenseval}
Marcos Zampieri, Shervin Malmasi, Preslav Nakov, Sara Rosenthal, Noura Farra,
  and Ritesh Kumar. 2019{\natexlab{b}}.
\newblock {SemEval-2019 Task 6: Identifying and Categorizing Offensive Language
  in Social Media (OffensEval)}.
\newblock In \emph{Proceedings of SemEval}.

\bibitem[{Zampieri et~al.(2020)Zampieri, Nakov, Rosenthal, Atanasova,
  Karadzhov, Mubarak, Derczynski, Pitenis, and
  {\c{C}}{\"o}ltekin}]{zampieri2020semeval}
Marcos Zampieri, Preslav Nakov, Sara Rosenthal, Pepa Atanasova, Georgi
  Karadzhov, Hamdy Mubarak, Leon Derczynski, Zeses Pitenis, and
  {\c{C}}a{\u{g}}r{\i} {\c{C}}{\"o}ltekin. 2020.
\newblock {SemEval-2020 Task 12: Multilingual Offensive Language Identification
  in Social Media (OffensEval 2020)}.
\newblock \emph{Proceedings of SemEval}.

\bibitem[{Zhang et~al.(2018)Zhang, Chang, Danescu-Niculescu-Mizil, Dixon, Hua,
  Taraborelli, and Thain}]{zhang2018conversations}
Justine Zhang, Jonathan Chang, Cristian Danescu-Niculescu-Mizil, Lucas Dixon,
  Yiqing Hua, Dario Taraborelli, and Nithum Thain. 2018.
\newblock Conversations gone awry: Detecting early signs of conversational
  failure.
\newblock In \emph{Proceedings of ACL}.

\end{thebibliography}
\bibliographystyle{acl_natbib}

\end{document}